%% file: arxiv.tex
\documentclass[10pt,twocolumn,letterpaper]{article}

\usepackage[pagenumbers]{cvpr} 


\definecolor{cvprblue}{rgb}{0.21,0.49,0.74}
\usepackage[pagebackref,breaklinks,colorlinks,allcolors=cvprblue]{hyperref}

\usepackage{colortbl}
\usepackage{xfrac}
\usepackage{setspace}
\usepackage{multirow}
\usepackage{amssymb}
\usepackage{pifont}
\usepackage{booktabs}       
\usepackage{amsfonts}       
\usepackage{nicefrac}       
\usepackage{microtype}

\usepackage{amsmath}

\usepackage{subfloat}

\usepackage{color}
\usepackage[dvipsnames]{xcolor}
\newcommand{\xmarkg}{\textcolor{lightgray}{\ding{55}}}%

\usepackage{datasheet}

\usepackage{color}
\usepackage[dvipsnames]{xcolor}

\def\paperID{1628} 
\def\confName{CVPR}
\def\confYear{2025}

\title{MSSIDD: A Benchmark for Multi-Sensor Denoising}

\author{
	Shibin Mei\textsuperscript{\rm 1}\footnotemark[1] \quad \quad 
	Hang Wang\textsuperscript{\rm 1}\thanks{Equal Contribution.} \quad \quad 
	Bingbing Ni\textsuperscript{\rm 1,2}\thanks{Corresponding author: Bingbing Ni.} 
        \\
	\textsuperscript{\rm 1}Huawei   \quad
	\textsuperscript{\rm 2}Shanghai Jiao Tong University 
   \\ \{shibin.mei1027, \, francis970625\}@gmail.com \,
	nibingbing@sjtu.edu.cn
}

\begin{document}
\maketitle

\begin{abstract}
The cameras equipped on mobile terminals employ different sensors in different photograph modes, and the transferability of raw domain denoising models between these sensors is significant but remains sufficient exploration. Industrial solutions either develop distinct training strategies and models for different sensors or ignore the differences between sensors and simply extend existing models to new sensors, which leads to tedious training or unsatisfactory performance. In this paper, we introduce a new benchmark, the Multi-Sensor SIDD (MSSIDD) dataset, which is the first raw-domain dataset designed to evaluate the sensor transferability of denoising models. The MSSIDD dataset consists of 60,000 raw images of six distinct sensors, derived through the degeneration of sRGB images via different camera sensor parameters. Furthermore, we propose a sensor consistency training framework that enables denoising models to learn the sensor-invariant features, thereby facilitating the generalization of the consistent model to unseen sensors. 
We evaluate previous arts on the newly proposed MSSIDD dataset, and the experimental results validate the effectiveness of our proposed method. 
Our dataset is available at \url{https://www.kaggle.com/datasets/sjtuwh/mssidd}.

\end{abstract}

\section{Introduction}
\label{sec_intro}

Optical signals from the physical world undergo a series of processes in cameras to obtain electrical signals (images), that conform to human eyes. Such a process is known as image signal processing (ISP), a core step of which is raw domain denoising. Recent years have witnessed notable progress in image restoration~\cite{dabov2007image,zhang2017beyond,zamir2020learning,zamir2021multi,zamir2022restormer,chen2022simple}, which has been widely applied to raw domain denoising in mobile terminals. Images from camera sensors usually contain shot and read noise, which exhibit significant differences across various lenses and sensors~\cite{brooks2019unprocessing,jin2023lighting}, and simply extending the denoising model to new sensors will result in unfavorable performance. Therefore, the transferability of raw domain denoising models across sensors becomes a non-negligible issue.


The memory space allocated to the camera for denoising in mobile terminals is limited, thus training different models for each specific sensor is not a wise choice. Moreover, different noise levels of sensors might require different training strategies, and optimizing for different sensors also entails a significant amount of development workload. It is common to merge some sensors with similar noise levels, using a single model during training to fit low-quality and high-quality data pairs from multiple sensors. However, directly merging data from these sensors often leads to performance degradation because even similar sensors may exhibit different characteristics (See Fig.~\ref{fig_noise_stat}), making the denoising model hard to optimize. Moreover, models directly merged are difficult to extend to new sensors, and fine-tuning the model with newly added sensors might damage the performance of previously optimized sensors. In this paper, we focus on enhancing the sensor transferability of denoising models, allowing the trained models to be easily transferred to unseen sensors without compromising denoising performance.

The sensor generalization relies on a multi-sensor dataset, which is currently lacking in the community. Existing datasets have little difference among different data sources and are not well partitioned to enable transferability assessment. To address this data scarcity problem and encourage future research, we introduce a multi-sensor SIDD (MSSIDD) dataset for the first time, a raw-domain benchmark designed for simulating multi-sensor scenarios. We refer to the raw domain processing of real cameras and inversely transform the sRGB images from the SIDD dataset to the raw domain~\cite{brooks2019unprocessing}. After the inverse transformation and mosaicing, we obtain clean raw images as captured by camera sensors. We collect calibration parameters from six sensors and add noise to the inversely transformed clean raw images according to these parameters, resulting in degenerated noisy images corresponding to each sensor. Totally, we obtain 60,000 noisy and clean raw image pairs for six different sensors.

\begin{figure*}[t]
\centering
\vspace{-2mm}
\includegraphics[width=0.96\textwidth]{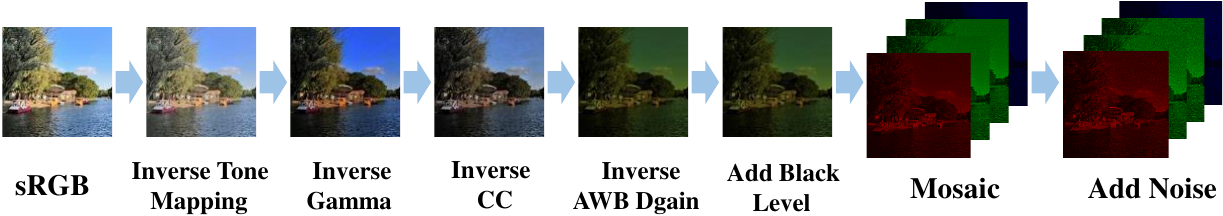}
\caption{The inverse transformation of ISP pipeline to obtain degenerated raw images.}
\label{fig_upi_exp} 
\end{figure*}

\begin{figure*}[t]
\vspace{-2mm}
\centering
\includegraphics[width=0.96\textwidth]{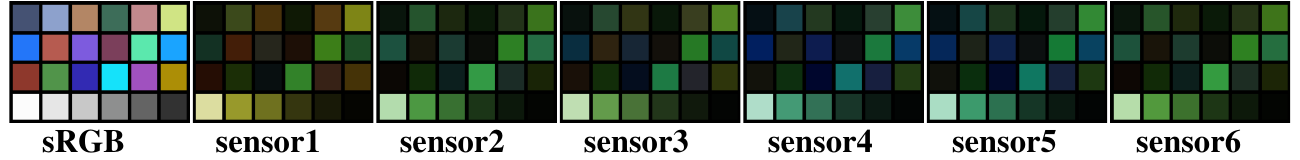}
\caption{The impact of white balance and color correction of different sensors on standard Macbeth color-checker.}
\vspace{-3mm}
\label{fig_colorckecker} 
\end{figure*}

Based on the multi-sensor dataset, we propose a sensor consistency method that encourages learning sensor-invariant representations during network training. For different sensors of the same image, we encourage the network to extract features that are as similar as possible, which we term the corresponding constraint as \textbf{intra-image supervision}. This approach can also be interpreted as encouraging the network to learn sensor-invariance denoising ability thereby decoupling the handling of sensor differences from the denoising. Then, we believe that such consistency supervision can also occur between different images. Specifically, we designate a portion of the network as primarily responsible for denoising, and we impose consistency constraints at the input and output positions of this sub-network, which we term \textbf{inter-image supervision}. We use the differences in the inputs of multiple images to the sub-network as a reference to constrain the differences in their corresponding outputs. Inspired by recent work in the domain generalization~\cite{ganin2015unsupervised}, we introduce adversarial training into our optimization. We design a sensor classification network, which increases sensor confusion through a gradient reversal layer, thereby forcing the network to extract sensor-independent features. These sensor consistency supervisions endow the network with excellent sensor transferability.

To establish baselines and verify the effectiveness of our proposed method, we evaluate several representative image-denoising methods on the MSSIDD dataset. The experimental results demonstrate that our method possesses effective sensor transfer capabilities.

Our contributions can be summarized as follows:
(1) We introduce the MSSIDD dataset, which includes data pairs generated by six camera sensors, for assessing the sensor transferability of denoising methods.
(2) We propose a sensor consistency supervision method through carefully designed intra-image and inter-image supervision combined with adversarial training, which imparts strong sensor generalization to denoising models.
(3) We evaluate several classic denoising methods to establish baselines and discuss the effectiveness of our method.

\section{Related Work}
\label{sec_related}

Image denoising aims to remove noise from images to obtain clean images and has made significant progress~\cite{zamir2020learning,wang2022uformer,chen2022simple, zamir2022restormer,chen2024intrinsic} in recent years.
Traditional methods~\cite{dabov2007image,buades2005non,gu2014weighted,osher2005iterative} usually employ hand-crafted image priors, which are flexible in handling denoising problems. 
CNN-based early pioneers~\cite{zhang2017beyond,zhang2017learning,zhang2018ffdnet,guo2019toward,gharbi2016deep} achieve impressive image restoration performance. To further improve the denoising performance, a flurry of architecture designs like skip connection~\cite{zhang2021plug}, U-shape structure~\cite{zamir2021multi}, attention mechanism~\cite{liu2018non,anwar2019real,zamir2020cycleisp} and others~\cite{jia2019focnet,zhang2023ingredient,cui2023focal} are widely explored.
Recently, transformer-based~\cite{vaswani2017attention,wang2023omni} denoising 
approaches~\cite{zamir2022restormer,liang2021swinir,wang2022uformer} have significantly advanced the performance by providing powerful capability of capturing long-range pixel dependencies. 

The evolution of image denoising also relies on the development of denoising datasets, such as DND~\cite{plotz2017benchmarking}, SIDD~\cite{abdelhamed2018high}. Based on the proposal of some inverse transformation methods~\cite{brooks2019unprocessing}, some denoising data generation methods~\cite{jin2023lighting, li2024efficient} have also been proposed for raw domain denoising in recent years. 
\textbf{Data generation in previous methods usually comes from different cameras, but they either have little difference among cameras or are not well partitioned, making it difficult to directly apply to the evaluation of the transferability of denoising models.} 
Based on this, we propose the first  multi-sensor dataset for evaluating the sensor .

\section{Dataset}
\label{sec_dataset}
Current learning-based denoising methods rely on large amount of paired  data~\cite{anwar2019real,zamir2020learning,zamir2021multi,wang2022uformer}, and directly collecting such data using cameras is difficult and laborious, since aligning images captured by different sensors is challenging. 
Previous works, such as~\cite{brooks2019unprocessing}, generate paired training data by starting from internet images and performing inverse transformations to obtain images before camera ISP post-processing. Noise is then added based on noise calibration parameters to simulate real camera noise on raw images. 
With such inspiration, to facilitate research on multi-sensor transferability, we develop the first multi-sensor denoising dataset from existing sRGB images.
Since the source clean sRGB images are derived from the ground truth of the SIDD dataset~\cite{abdelhamed2018high}, we name our multi-sensor denoising dataset the Multi-Sensor SIDD (MSSIDD).

\subsection{ISP Pipeline with Inverse}
\label{subsec_isp}

The distribution inherent in the electrical signals output by a camera optical sensor differs significantly from the image distribution of the natural world. This necessitates a series of post-processing steps for converting the captured raw images into the sRGB domain. In this section, we provide a concise overview of the steps involved in transforming the captured electrical signals into sRGB images that align with human vision. To facilitate the inverse transformation of sRGB images, we will also describe the reverse processes of some of these operations.

\textbf{Noise.} The noise in raw images can be broadly categorized into multiplicative noise and additive noise~\cite{boie1992analysis}. Multiplicative noise arises from fluctuations in the number of photons received by the camera's optical sensor pixels per unit time. Such noise, akin to the mottled rain marks on the ground during a rainy day, 
is known as photon shot noise~\cite{beenakker1999photon} and follows a Poisson distribution.
Additive noise originates from the voltage fluctuations during signal processing, including signal readout, ISO gain, and analog-to-digital conversion. This noise, known as read noise~\cite{chang2020learning}, follows a zero-mean Gaussian distribution. By combining these two noise sources, we can establish the relationship between the noisy sensor output $I$ and real signal (i.e., Analog-Digital Units, ADU) $x$ as, 
$I \sim \mathcal{N}(x, \sigma^2_{shot}x+\sigma^2_{read})$,
where $\sigma^2_{read}$ is the variance of read noise, and $\sigma^2_{shot}x$ is the variance of photon shot noise.
Subsequently, we formulate shot noise and read noise as functions of camera light sensitivity, i.e., ISO,
\begin{equation}
\label{eq_read_shot}
\begin{split}
&\sigma^2_{shot} = K_0 \times ISO + K_1, \\
&\sigma^2_{read} = B_0 \times ISO^2 + B_1 \times ISO + B_2 
\end{split}
\end{equation}
where the camera noise parameters $K_0,K_1,B_0,B_1,B_2$ can be obtained through calibration~\cite{healey1994radiometric}.

\textbf{Denoising and Demosaicing.}  Due to the simple noise distribution in raw image~\cite{wang2020practical}, ISP systems typically perform denoising on the raw domain to obtain clean raw image. The RGB components of the raw image are arranged in a Bayer pattern~\cite{ramanath2002demosaicking}, which mirrors the pixel arrangement in the camera optical sensor. 
To produce a full-resolution RGB image, various demosaicing algorithms~\cite{getreuer2011malvar,buades2009self,hirakawa2005adaptive,pekkucuksen2010gradient} have been developed. We employ bi-linear interpolation for demosaicing, following~\cite{brooks2019unprocessing}. The inverse process of demosaicing, known as mosaicing, is straightforward. 
We simply down-sample the corresponding channels of the RGB image to obtain the 4-channel raw image.

\textbf{Digital Gain.} 
Current ISP adjusts the brightness of images by uniformly modifying the values of all pixels in response to overexposure or underexposure in the raw image, a process handled by the camera auto-exposure (AE) algorithm~\cite{bernacki2020automatic}. 
Since the auto-exposure algorithm functions as a black box, it is challenging to determine the exact digital gain (dgain) value for each image. In this study, we set a fixed dgain value for simplicity. The inverse process involves multiplying each pixel value by the reciprocal of dgain, defined as $dgain_{inverse} = dgain^{-1}$. 
To ensure diversity in our dataset, we randomly sample the inverse gain value for each image, where we assume the inverse gain follows a Gaussian distribution with 0.65 mean and 0.2 standard deviation.

\begin{figure*}[!t]
\centering
\includegraphics[width=1.0\textwidth]{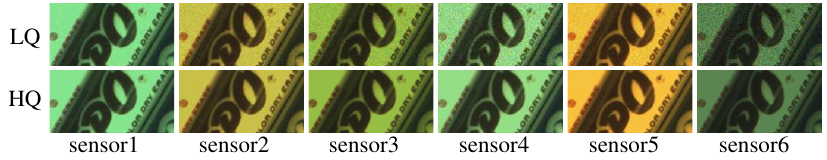}
\caption{Noisy and clean image pairs of six sensors. Demosaicing is applied for better visualization.}
\label{fig_image_exp} 
\end{figure*}

\textbf{White Balance.} 
White balance is the process of removing unrealistic color casts so that objects that appear white in person are rendered white in the photo. 
Typical white balance~\cite{barron2015convolutional,afifi2021cross} is achieved by adjusting the digital gain of the red and blue channels. For a specific camera sensor, we can obtain its white balance parameters under different standard light sources, represented by the red gains $r_{gain}$ and blue gains $b_{gain}$. Common standard light sources~\cite{wiki:standard_illuminant} include D65, D75, D50, TL84, CWF, U35, 
and we form a convex combination of the white balance parameters from multiple light sources, as real-life ambient lighting often involves a mix of radiation from different sources~\cite{li2024efficient}. We randomly select the AWB parameters of two light sources and perform a random weighted average to obtain the AWB parameters $[ r_{gain},1,b_{gain} ]$.
The inverse AWB process involves taking the reciprocal of the AWB parameters and applying the corresponding channel gains. 
To prevent the synthetic dataset from lacking highlights, we followed~\cite{brooks2019unprocessing} to safely handle saturated pixels.

\textbf{Color Correction.} 
The color characteristics of camera sensors usually do not satisfy the Luther condition~\cite{heuser2012human}, meaning the sensor RGB responses are not linearly independent. 
Therefore, color correction is used for correcting the sensor color characteristics to approximate a standard observer~\cite{heuser2012human}. For simplicity, such a process is often represented as a 3$\times$3 matrix, known as the color correction matrix (CCM). Camera manufacturers typically provide preset color correction matrices for daylight and nighttime~\cite{li2024efficient}, referred to as $CCM_d$ and $CCM_n$. During data generation, we randomly select two light sources and perform a random linear combination of their corresponding CCMs,
$CCM = \alpha CCM_d + (1 - \alpha) CCM_n$.
The inverse process involves applying the inverse of the CCM to the pixel values, given by $CCM_{inverse}=CCM^{-1}$.

\textbf{Gamma Correction.} Gamma correction allocates more bits of dynamic range to low-intensity pixels, as human perception is more sensitive to gradients in darker areas of an image~\cite{guo2004gamma}. Gamma correction and its inverse are applied as,
$f(x) = x^{\gamma},f^{-1}(y) = y^{1/\gamma}$,
where $\gamma$ is usually set as $1/2.2$.

\textbf{Tone Mapping.} Following~\cite{brooks2019unprocessing}, we perform the inverse tone mapping for image generation, i.e.,
\begin{equation}
TM^{-1}(y)=\frac{1}{2}-\sin(\frac{\sin^{-1}(1-2y)}{3})
\end{equation}

\subsection{Data Generation}
\label{subsec_datagen}

The electrical signals captured by the camera optical sensor undergo a series of ISP processes to generate sRGB images that conform to human visual perception. This process typically includes raw domain denoising, demosaicing, digital gain, white balance, color correction, gamma correction, and tone mapping. Starting from clean sRGB images from the SIDD dataset, we sequentially apply inverse tone mapping, inverse gamma correction, inverse color correction, inverse white balance with digital gain adjustments, and inverse demosaicing (mosaic), to simulate the clean images in the raw domain, as shown in Fig.~\ref{fig_upi_exp}. We display the impact of white balance and color correction on image color, as shown in Fig.~\ref{fig_colorckecker}.

With camera parameters of six sensors collected from ~\cite{sonysensor}, for each image from SIDD, we generate six corresponding clean raw images following the BGGR Bayer pattern tailored to each sensor. 
We then induce noise degradation on the clean raw images for specified ISO values to obtain noisy raw images. Specifically, we first subtract the black level (BLC) from the clean raw images to mitigate the influence of dark current, and then randomly select an ISO value from the range of 2400$\sim$12800. We compute the intensities of photon shot noise and read noise as Eqn.\ref{eq_read_shot}, thus obtaining the variance of noise as $\sigma^2 = \sigma^2_{shot} x + \sigma^2_{read}$. The random noise is then generated following distribution $\mathcal{N}\sim(0, \sigma^2)$, and we finally overlay it onto the original clean raw image. After adding the black level compensation, we obtain the target noisy raw image, as shown in Fig.~\ref{fig_image_exp}, where we can observe significant distinctions among different sensors.

\begin{figure*}[!t]
    \centering
    \subfloat[SNR(DB)]
    {\includegraphics[width=0.32\linewidth]{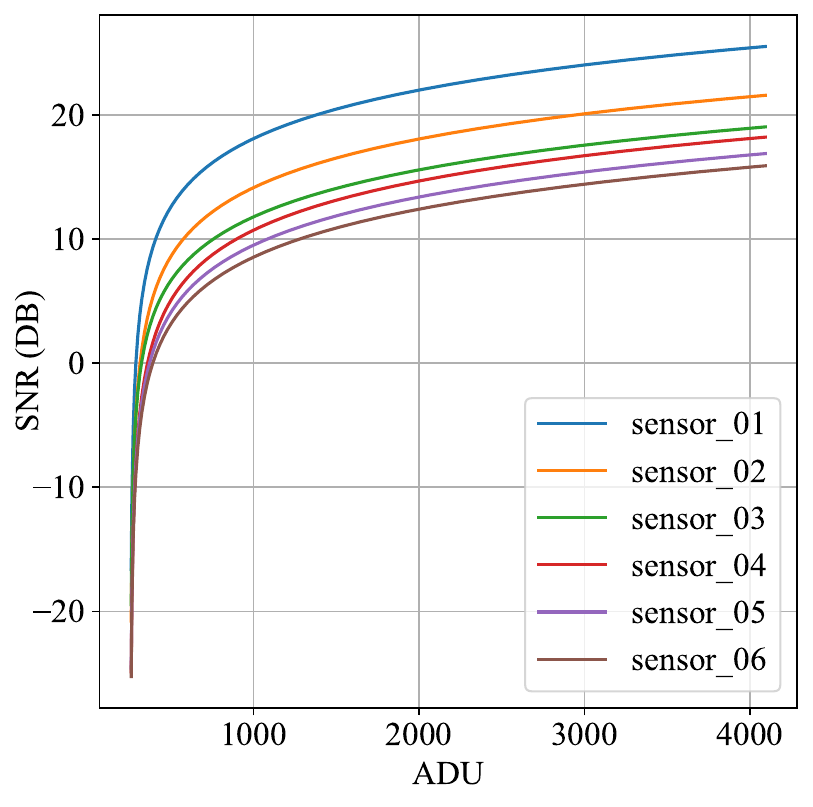}}
    \subfloat[Noise Variance]
    {\includegraphics[width=0.32\linewidth]{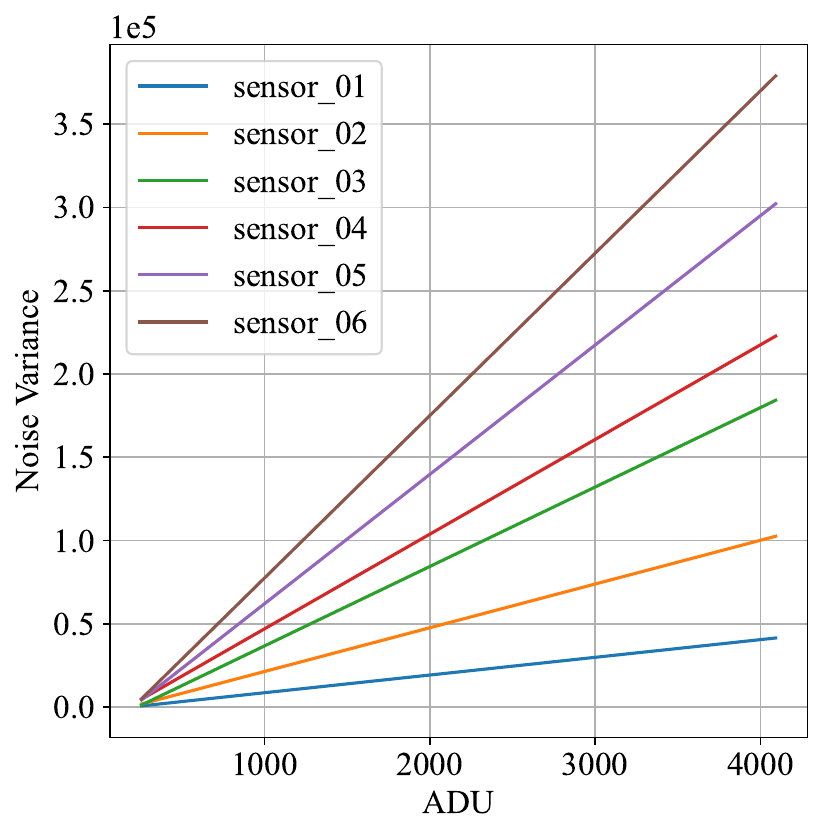}}
    \subfloat[Shot and Read Noise]
    {\includegraphics[width=0.35\linewidth]{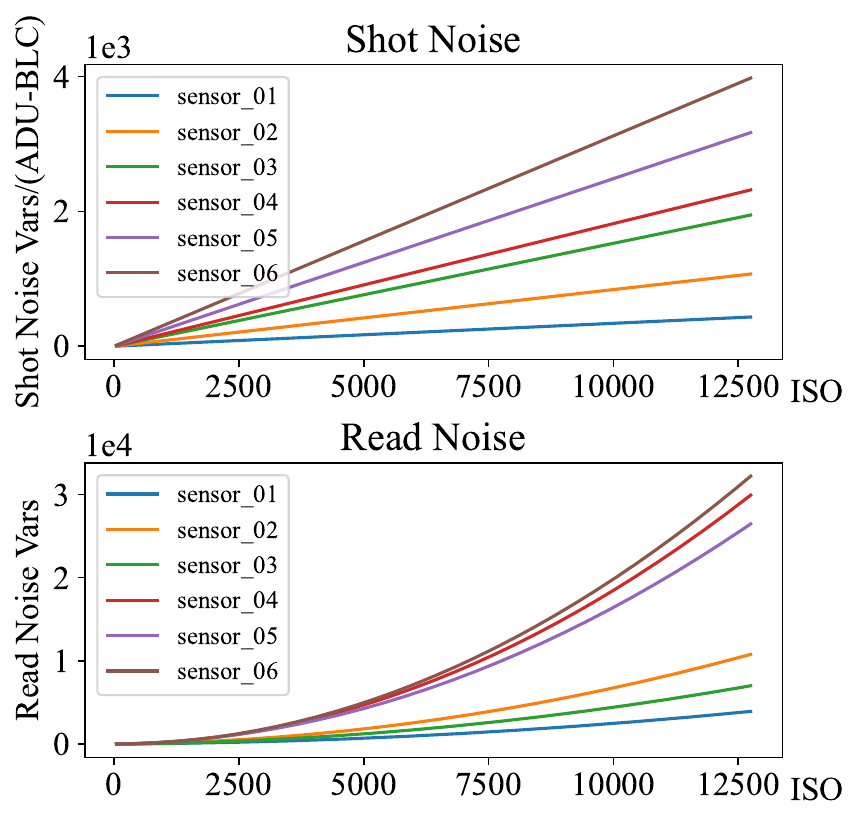}}
\caption{Analysis of noise statistic of six sensors. We display the SNR and total noise variance concerning ADU. We also present the relationship between read and shot noise and camera ISO.}
\label{fig_noise_stat}
\end{figure*}

\subsection{Benchmark}
\label{subsec_benchmark}

We extract 10,000 images from the training set and 1,000 images from the test set of the SIDD dataset~\cite{brooks2019unprocessing}, and generate the MSSIDD dataset after degradation and noise addition. The MSSIDD dataset comprises six sensors denoted as sensor1, sensor2, sensor3, sensor4, sensor5, and sensor6, resulting in a total of 60,000 training data pairs and 6,000 validation data pairs. For each image in the dataset, we also recorded the dgain, AWB, and CCM parameters, which are stored in the file $meta\_data.pkl$. 
We analyze the SNR and total noise variance concerning real signal ADU for each sensor, and the read and shot noise with respect to camera ISO, as shown in Fig.\ref{fig_noise_stat}.
We can observe that the six sensors of the MSSIDD dataset cover varying degrees of noise intensity.

The MSSIDD dataset can be used for research on sensor transferability. Given a set of known sensors $M_{known}$ and an unknown newly added sensor $M_{new}$, the objective is to generalize the model trained on the known sensors to the newly added sensor. The model performance is evaluated on the newly added sensor $M_{new}$. For a known sensor $m$ within $M_{known}$, the data pairs can be denoted as $(X^m,Y^m)$. Considering a denoising network $F_{\theta}:X\rightarrow Y$, where $\theta$ represents the network parameters, we define the loss on this sensor as:
\begin{equation}
R^m(\theta)=\mathbb{E}[\mathcal{L}(F_{\theta}(X^m),Y^m)],
\end{equation}
where $\mathcal{L}$ denotes reconstruction loss, such as PSNR loss. The objective of sensor transferring is to minimize the reconstruction loss on the worst sensor among all sensors,
\begin{equation}
\mathop{\min}_{\theta\in\Theta} \mathop{\max}_{m\in M} R^m(\theta).
\end{equation}

Under the limit on the number of denoising models on mobile terminals, we can not customize each sensor with a specific model. Fine-tuning on newly added sensors may damage the performance of known sensors, and adapting existing models to new sensors is a better choice.
Considering the domain shift between $M_{known}$ and $M_{new}$, sensor generalization is an important and challenging problem. 
It should be noted that our datasets can be used for unsupervised and semi-supervised tasks.

Our MSSIDD dataset will be publicly available at \url{https://www.kaggle.com/datasets/sjtuwh/mssidd} and licensed under the Apache License, Version 2.0, January 2004. 

\section{Method}
\label{sec_method}

In this section, we elaborate on sensor consistency supervision, which is achieved through intra- and inter-image constraints, coupled with adversarial training, to endow the denoising model with robust sensor generalization capabilities. The proposed supervision encourages the model to decouple the processing of sensor-related information from denoising functions, enabling certain sub-structures of the model to focus solely on denoising tasks, which facilitates excellent transferability. Our model structure can be seen in Fig.\ref{fig_network}.

\subsection{Intra- and Inter-Image Supervision}
\label{subsec_interintra}

\begin{figure}
\centering
\vspace{-0.5cm}
\includegraphics[width=0.6\linewidth]{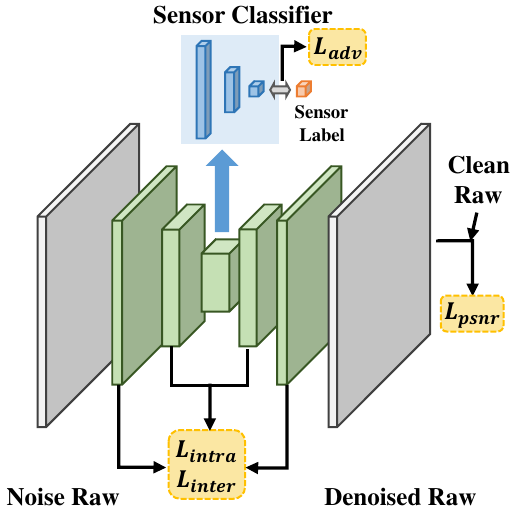}
\caption{Model Structure. Intra- and inter-image sensor consistency supervision and adversarial learning are equipped to facilitate sensor generalization.}
\label{fig_network}
\end{figure}
Assuming the denoising network as $F_{\theta}$, with a subnetwork (green part in Fig~\ref{fig_network}) extracted from $F_{\theta}$ as $F^{sub}_{\theta}$, we expect the subnetwork to solely focus on denoising, effectively extracting sensor-agnostic features. Denoising networks typically entail downsampling and upsampling of feature maps, designed with approximate symmetric structures, often incorporating multi-scale network architectures such as UNet~\cite{unet}. Consequently, it is convenient to extract consistent-sized features from symmetric positions within the subnetwork.
For an image $I$, we can obtain the corresponding noisy images for $N$ sensors. 
At the $l$-th stage of the subnetwork, we extract scale-consistent inputs and outputs, denoted as $U=\{U^l_1(I),U^l_2(I),...,U^l_N(I)\}$ and $V=\{V^l_1(I),V^l_2(I),...,V^l_N(I)\}$. Inspired by~\cite{sagawa2019distributionally, chen2020simple, mei2022towards}, we hope that the features generated by different sensors of the same image are as similar as possible, that is, features in $U$ or features in $V$ are as consistent as possible. However, there are also differences between noisy images of different sensors of the same image, and directly constraining the $U$ or $V$ may not be conducive to model convergence. Therefore, we design the supervision between different sensors of the same image in a relation-based form,
\begin{equation}
\begin{split}
\mathcal{L}^l_{intra} =  \frac{N(N-1)}{2 | \mathcal{D}|} & \sum_{I\in\mathcal{D}} \sum_{i,j} \| (V^l_i(I)-V^l_j(I)) \\
&- (U^l_i(I)-U^l_j(I)) \|
\end{split}
\end{equation}
where $\mathcal{D}$ denotes the whole dataset, and $| \mathcal{D} |$ denotes the number of images. $i, j$ denotes two sensors. In our experiment, we apply consistency constraints at multiple scales, as,
\begin{equation}
\begin{split}
\mathcal{L}_{intra} = \frac{N(N-1)}{2 | \mathcal{D}|L} \sum_{l} & \sum_{I\in\mathcal{D}} \sum_{i,j} \| (V^l_i(I)-V^l_j(I)) \\
&- (U^l_i(I)-U^l_j(I)) \|.
\end{split}
\end{equation}
We also posit that consistency constraints should exist between different images. For two images, we consider these two images as a naive video with only two frames, so we can constrain the sensor consistency of this video through relation-based consistency,
\begin{equation}
\small
\begin{split}
\mathcal{L}_{inter} = & \frac{N(N-1)|\mathcal{B}|(|\mathcal{B}|-1)}{4L}  \sum_{l} \sum_{I_m,I_n\in\mathcal{B}} \sum_{i,j} \| (V^l_i(I_m)  \\
&-V^l_j(I_n)) - (U^l_i(I_m)-U^l_j(I_n)) \|,
\end{split}
\end{equation}
where $\mathcal{B}$ denotes a data batch, and $| \mathcal{B} |$ denotes the batch size. We refer to $\mathcal{L}_{intra}$ and $\mathcal{L}_{inter}$ as \textbf{multi-scale multi-sensor consistency supervision}.

\subsection{Adversarial Training}
\label{subsec_advtrain}

We posit that features from the subnetwork $F^{sub}_{\theta}$ should contain as little sensor-specific information as possible. Inspired by~\cite{ganin2015unsupervised, arjovsky2019invariant, nam2019reducing, xu2022graphical, zhao2023variational, mei2023exploring}, we perform sensor classification on the features output by the subnetwork and maximize the classification loss.

Assume the feature $f$ is obtained through feature extraction network $F_{\theta}(I)$, then the network branches into two paths to process the feature. One path goes through the reconstruction network $F_r(f,\theta_r)$ for image reconstruction, resulting in denoised output. The other path goes through the sensor classification network $F_c(f,\theta_c)$ for sensor classification. For $N$ sensors, the model $F_c$ acts as an N-class classification network. During training, we aim to minimize the reconstruction loss $\mathcal{L}_r$ for denoising and simultaneously maximize confusion among features from different sensors to maximize sensor classification loss $\mathcal{L}_c$. The adversarial training loss can be represented as,
\begin{equation}
\small
\mathcal{L}_{adv} = \mathbb{E}_{X_i} [ \mathcal{L}_r (F_r(\theta_r, X_i),Y_i) - \alpha \mathcal{L}_c (F_c(\theta_c, X_i), Y_{ic}) ]
\end{equation}
where $(X_i, Y_i)$ is the data pair, and $Y_{ic}$ is the corresponding sensor label. $\alpha$ denotes the balance hyperparameter. To enable end-to-end training, we incorporate a gradient reversal layer, following~\cite{ganin2015unsupervised}. 

In summary, the final sensor consistency supervision is defined as
\begin{equation}
\mathcal{L}_{MS} =\lambda_{1} \mathcal{L}_{inter} + \lambda_{2} \mathcal{L}_{intra} + \lambda_{3} \mathcal{L}_{adv},
\end{equation}
where $\lambda_{1}, \lambda_{2}, \lambda_{3}$ are balance weights.

\section{Benchmark Experiments}
\label{sec_exp}

In this section, we evaluate several representative denoising methods on our MSSIDD dataset. We also verify that sensor transferability can be improved with the designed consistency supervision.

\subsection{Experimental Setup} 
\label{subsec_expsetup}

\textbf{Implementation details.} 
Since most previous image denoising methods are applied on RGB images with 3 channels, these methods can not be directly evaluated on raw images with 4 channels. We re-implement these methods with their open-source code released on the GitHub platform, and adjust the dimension of the input channel from 3 to 4 to adapt for raw domain data. For a fair comparison, all methods are trained with the same experimental settings. Specifically, we use a standard PSNR loss with the AdamW ~\cite{adamw} optimizer to train the model for 100K iterations on two NVIDIA V100 GPUs. We set the batch size as 2 for each sensor on each GPU, resulting in a total batch size of $2\times Sensor\_Num$ per GPU. 
The learning rate is initially set to $1e-3$, and then follows a cosine decay to the minimum learning rate of $1e-7$. We randomly crop image patches as $128 \times 128$ as input during training.
The balance parameters $\lambda_{1}$, $\lambda_{2}$, $\lambda_{3}$ are set as 0.1, 0.1, 1.0 respectively. We set $\alpha$ from 0 to 1 as training iterations increase.
In all experiments, no other data augmentation strategy is adopted.
Our implementation is publicly available at Github~\footnote{https://github.com/shibin1027/MSSIDD}. 

\begin{table*}[t]
\centering
\small
\begin{spacing}{1.0}
\setlength{\tabcolsep}{3.2mm}
\begin{tabular}{l|c|c|c|c|c|c }
\toprule
\multicolumn{1}{c|}{} & 
$\rightarrow$ sensor1 & 
$\rightarrow$ sensor2 & $\rightarrow$ sensor3 & $\rightarrow$ sensor4 & $\rightarrow$ sensor5 & $\rightarrow$ sensor6 \\ 
Algorithm &  PSNR / SSIM &  PSNR / SSIM &  PSNR / SSIM &  PSNR / SSIM &  PSNR / SSIM &  PSNR / SSIM \\ \hline
BM3D~\cite{dabov2007image} & 41.59 / 0.9344 & 41.08 / 0.9326 & 41.09 / 0.9308 & 39.88 / 0.9273 & 40.66 / 0.9339 & 39.87 / 0.9294\\
DnCNN~\cite{zhang2017beyond} & 44.18 / 0.9710 & 43.66 / 0.9672 & 43.23 / 0.9652 & 41.35 / 0.9549 & 42.87 / 0.9648 & 41.23 / 0.9538\\ 
FFDNet~\cite{zhang2018ffdnet} & 44.51 / 0.9707 & 44.23 / 0.9688 & 43.93 / 0.9675 & 42.50 / 0.9623 & 43.61 / 0.9679 & 41.92 / 0.9587\\ 
RIDNet~\cite{anwar2019real} & 45.76 / 0.9763 & 44.93 / 0.9727 & 44.85 / 0.9724 & 43.36 / 0.9668 & 44.28 / 0.9715 & 43.30 / 0.9668\\ 
AINDNet~\cite{kim2020transfer} & 45.82 / 0.9769 & 45.03 / 0.9732 & 44.99/ 0.9729 & 43.50 / 0.9670 & 44.39 / 0.9720 & 43.42 / 0.9668\\
CycleISP~\cite{zamir2020cycleisp} & 45.95 / 0.9772 & 45.28 / 0.9739 & 45.12 / 0.9733 & 43.64 / 0.9678 & 44.56 / 0.9727 & 43.56 / 0.9673\\
MIRNet~\cite{Zamir2020MIRNet} & 46.07 / 0.9774 & 45.44 / 0.9740 & 45.20 / 0.9736 & 43.88 / 0.9689 & 44.75 / 0.9734 & 43.70 / 0.9682\\
MPRNET~\cite{zamir2021multi} & 46.11 / 0.9774 & 45.28 / 0.9741 & 45.20 / 0.9737 & 43.86 / 0.9689 & 44.80 / 0.9736 & 43.74 / 0.9685\\
HINET~\cite{chen2021hinet}  & 46.22 / 0.9775 & 45.48 / 0.9744 & 45.42 / 0.9741 & 44.19 / 0.9697 & 45.10 / 0.9741 & 44.10 / 0.9694\\
Uformer~\cite{wang2022uformer} &  46.32 / 0.9778  &   45.55 / 0.9747   & 45.49 / 0.9743 & 44.26 / 0.9700 & 45.21 / 0.9744 & 44.17 / 0.9696\\
NAFNet~\cite{chen2022simple} & 46.41 / 0.9783 & 45.66 / 0.9753 & 45.58 / 0.9749 & 44.38 / 0.9708 & 45.30 / 0.9751 & 44.29 / 0.9704\\ 
Restormer~\cite{zamir2022restormer} & 46.55 / 0.9789 & 45.81 / 0.9760 & 45.75 / 0.9756 & 44.55 / 0.9717 & 45.48 / 0.9760 & 44.45 / 0.9713\\
\hline
\rowcolor{green!10}\textbf{MS-NAFNet} & 46.54 / 0.9788 & 45.78 / 0.9758 & 45.66 / 0.9756 & 44.49 / 0.9716 & 45.44 / 0.9762 & 44.47 / 0.9715 \\ 
\rowcolor{green!10}\textbf{MS-Restormer} & 46.68 / 0.9795 & 45.95 / 0.9768 & 45.87 / 0.9762 & 44.72 / 0.9725 & 45.61 / 0.9767 & 44.56 / 0.9720 \\ 
\bottomrule
\end{tabular}
\end{spacing}
\caption{Performance comparison on the proposed MSSIDD Dataset under Raw2Raw setting.}
\label{tab:results_raw}
\end{table*}

\begin{figure*}[t]  
    \centering  
    \includegraphics[width=1.0\textwidth]{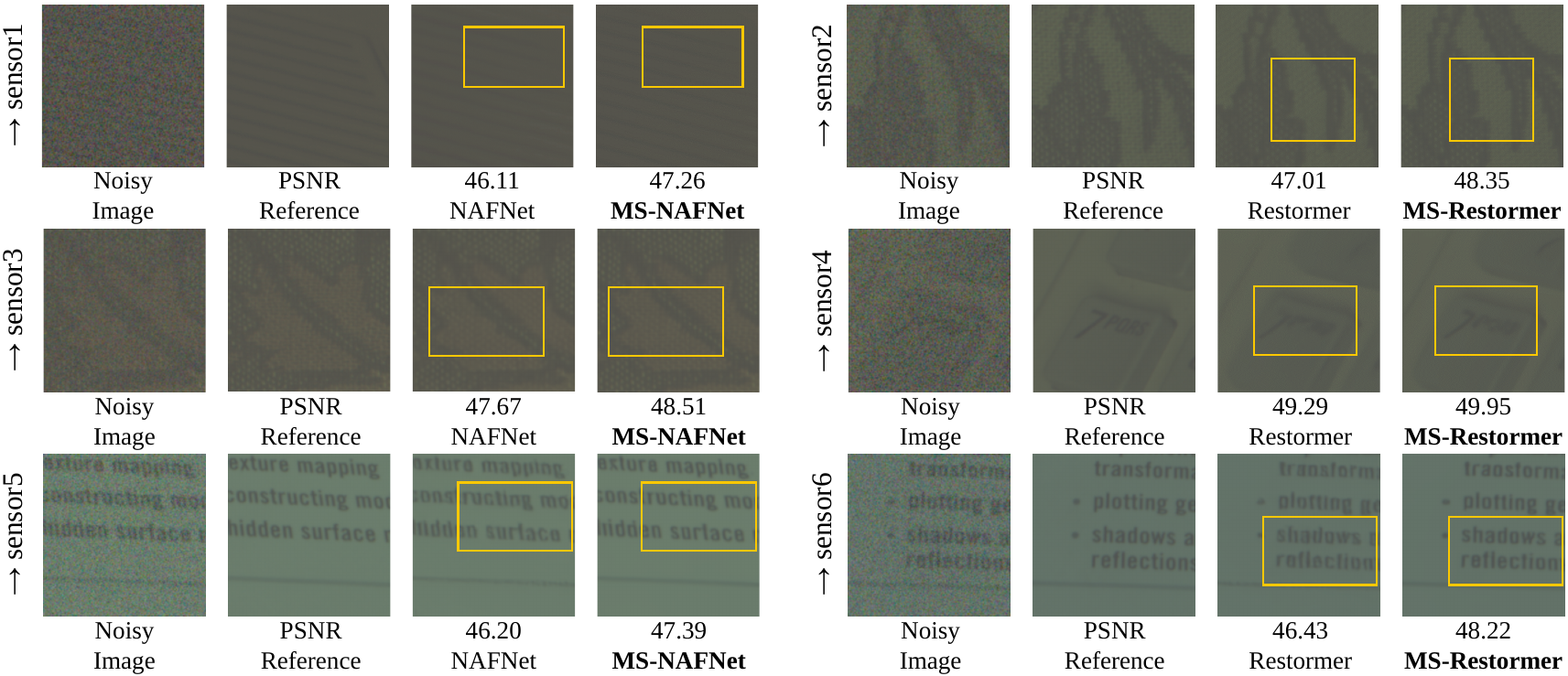}  
    \caption{Visual comparison of different methods under Raw2Raw setting on the proposed MSSIDD dataset. (Best viewed by zooming.)
    }  
    \label{fig_results}  
\end{figure*}

\begin{table*}[t]
\centering
\small
\begin{spacing}{1.0}
\setlength{\tabcolsep}{3.2mm}
\begin{tabular}{l|c|c|c|c|c|c }
\toprule
\multicolumn{1}{c|}{} & 
$\rightarrow$ sensor1 & 
$\rightarrow$ sensor2 & $\rightarrow$ sensor3 & $\rightarrow$ sensor4 & $\rightarrow$ sensor5 & $\rightarrow$ sensor6 \\ 
Algorithm &  PSNR / SSIM &  PSNR / SSIM &  PSNR / SSIM &  PSNR / SSIM &  PSNR / SSIM &  PSNR / SSIM \\ \hline
BM3D~\cite{dabov2007image} & 39.24 / 0.9345 & 38.85 / 0.9359 & 38.41 / 0.9340 & 37.51 / 0.9294 & 37.94 / 0.9328 & 36.72 / 0.9282\\
DnCNN~\cite{zhang2017beyond} & 41.16 / 0.9532 & 40.03 / 0.9536 & 40.27 / 0.9542 & 37.73 / 0.9450 & 38.69 / 0.9511 & 37.26 / 0.9367\\ 
FFDNet~\cite{zhang2018ffdnet} & 42.20 / 0.9577 &  41.47 / 0.9590  &  41.56 / 0.9581  & 40.49 / 0.9641 & 40.63 / 0.9596 &  39.73 / 0.9531 \\ 
RIDNet~\cite{anwar2019real} & 43.28 / 0.9642 & 42.25 / 0.9628 & 42.40 / 0.9624 & 40.58 / 0.9608 & 41.44 / 0.9632 & 40.31 / 0.9561\\ 
AINDNet~\cite{kim2020transfer} & 43.35 / 0.9641 & 42.41 / 0.9633 & 42.58 / 0.9627 & 40.81 / 0.9608 & 41.78 / 0.9634 & 40.67 / 0.9558\\
CycleISP~\cite{zamir2020cycleisp} & 43.60 / 0.9652 & 42.63 / 0.9640 & 42.71 / 0.9633 & 41.22 / 0.9622 & 42.07 / 0.9644 & 41.12 / 0.9579\\
MIRNet~\cite{Zamir2020MIRNet} & 43.81 / 0.9658 & 43.01 / 0.9652 & 43.21 / 0.9653 & 41.62 / 0.9638 & 42.37 / 0.9660 & 41.45 / 0.9596\\ 
MPRNET~\cite{zamir2021multi} & 43.88 / 0.9660 & 43.08 / 0.9653 & 43.25 / 0.9653 & 41.72 / 0.9641 & 42.47 / 0.9664 & 41.52 / 0.9602\\
HINET~\cite{chen2021hinet}  & 44.02 / 0.9661 & 43.27 / 0.9657 & 43.47 / 0.9654 & 41.88 / 0.9645 & 42.78 / 0.9670 & 41.76 / 0.9605\\
Uformer~\cite{wang2022uformer} &  44.18 / 0.9664  &  43.39 / 0.9660   & 43.53 / 0.9665 & 41.96 / 0.9646  & 42.81 / 0.9670 & 41.78 / 0.9604 \\
NAFNet~\cite{chen2022simple} & 44.33 / 0.9671 & 43.58 / 0.9669 & 43.72 / 0.9663 & 42.17 / 0.9657 & 42.99 / 0.9680 & 42.02 / 0.9617 \\
Restormer~\cite{zamir2022restormer} &  44.45 / 0.9679  &  43.78 / 0.9678  &  43.88 / 0.9671  & 42.32 / 0.9665 & 43.22 / 0.9690 & 42.16 / 0.9625\\
\hline
\rowcolor{green!10}\textbf{MS-NAFNet}    & 44.48 / 0.9678    & 43.70 / 0.9676 & 43.89 / 0.9669 & 42.30 / 0.9665 & 43.15 / 0.9689 & 42.19 / 0.9623 \\ 
\rowcolor{green!10}\textbf{MS-Restormer} &  44.57 / 0.9685  &  43.89 / 0.9687  &  44.02 / 0.9676  & 42.45 / 0.9672 & 43.36 / 0.9695 & 42.30 / 0.9631\\ 
\bottomrule
\end{tabular}
\end{spacing}
\caption{Performance comparison on the proposed MSSIDD Dataset under Raw2RGB setting.}
\label{tab:results_rgb}
\end{table*}

\begin{figure*}[t]
\begin{minipage}{0.56\linewidth}
\centerline{\includegraphics[width=7.1cm,height=
3.0cm]{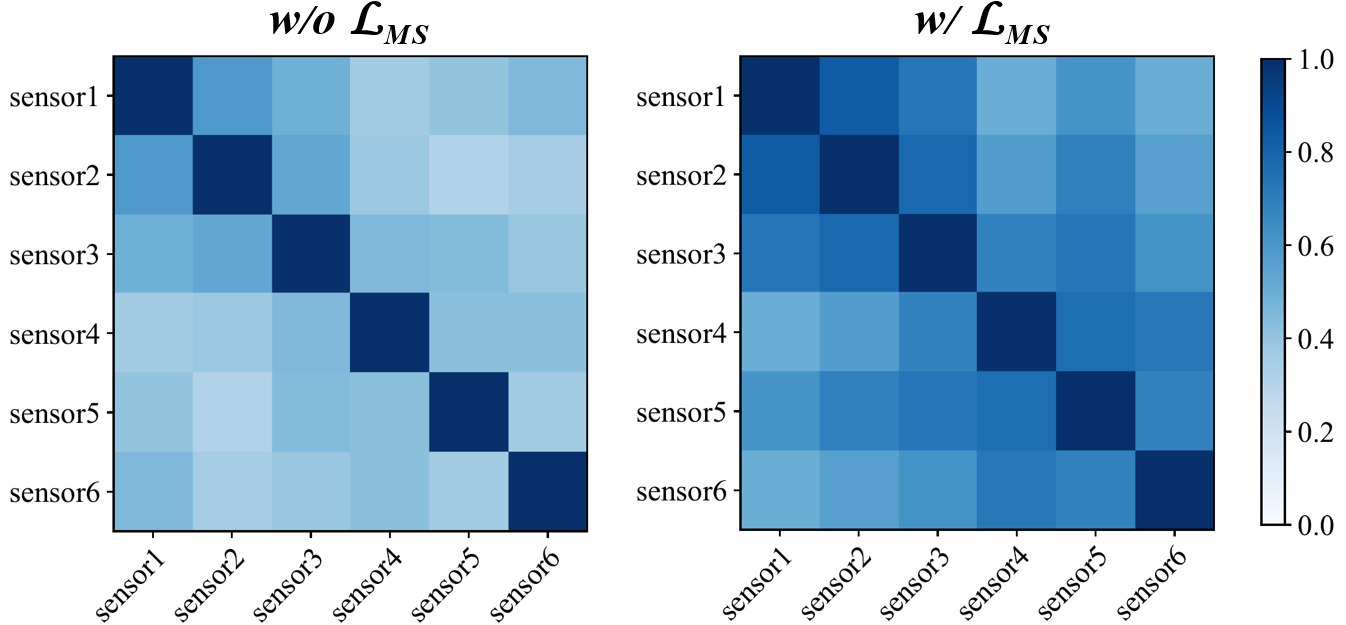}}
\end{minipage}
\hfill
\begin{minipage}{0.40\linewidth}
\centering
\small

\setlength\tabcolsep{3.0pt}
		\begin{tabular}{ccc|c}
			\toprule
        $\mathcal{L}_{inter}$ & $\mathcal{L}_{intra}$ & $\mathcal{L}_{adv}$ & PSNR / SSIM  \\
			\hline
	\xmarkg & \xmarkg  & \xmarkg             & 44.29 / 0.9704 \\
	 $\checkmark$ &   \xmarkg  &      \xmarkg   &  44.40 / 0.9711 \\
	 \xmarkg  &  $\checkmark$      & \xmarkg    &  44.36 / 0.9707 \\
	  \xmarkg  &   \xmarkg  &  $\checkmark$    & 44.38 / 0.9708 \\
	 $\checkmark$ & $\checkmark$ & $\checkmark$ & 44.47 / 0.9715 \\
			\bottomrule		
	\end{tabular}

\end{minipage}
\caption{Visualization of feature similarity across six sensors (left) and ablation studies of MS-NAF on ``$\rightarrow$ sensor6'' task under the Raw2Raw setting (right).}
\label{tab_fig: analysis}
\end{figure*}

\textbf{Performance comparison.} We evaluate several state-of-the-art methods on our MSSIDD benchmark. CNN based denoising methods, \emph{e.g.}, DnCNN~\cite{zhang2017beyond}, FFDNet~\cite{zhang2018ffdnet}, RIDNet~\cite{anwar2019real}, CycleISP~\cite{zamir2020cycleisp}, HINet~\cite{chen2021hinet}, NAFNet~\cite{chen2022simple}, and transformer based methods, \emph{e.g.} Uformer~\cite{wang2022uformer}, Restormer~\cite{zamir2022restormer}, are introduced for comparison.
We design two evaluation settings, \emph{Raw2Raw} and \emph{Raw2RGB}, on the proposed MSSIDD dataset. 
For the Raw2Raw setting, the supervision is directly applied to the raw domain, i.e., the output of our network. For the Raw2RGB setting, the denoised raw image will be transformed to the sRGB domain through an ISP pipeline~\cite{brooks2019unprocessing}, and we subsequently utilize the transformed sRGB image to calculate reconstruction loss.
We integrate our proposed supervision $\mathcal{L}_{MS}$ into NAFNet and Restormer, and derive two models named MS-NAFNet and MS-Restormer respectively, where `MS' denotes the multi-sensor consistency supervision.

\textbf{Evaluation Metric.} Following prior works~\cite{plotz2017benchmarking,abdelhamed2018high,timofte2017ntire}, we report peak signal-to-noise ratio (PSNR) and structural similarity (SSIM)~\cite{image_quality} as the evaluation metric on raw and sRGB outputs. 

More details can be found in the supplementary materials.

\subsection{Evaluation Results} 
\label{subsec_expresults}
\textbf{Quantitative Evaluation.} 
Table~\ref{tab:results_raw} benchmarks different image denoising methods on six sub-tasks under the Raw2Raw setting on the proposed MSSIDD dataset.
``$\rightarrow $sensor $m$" denotes the sub-task of transferring from other sensors to sensor $m$, where the data of sensor $m$ is unavailable during training.
It can be observed that both MS-NAFNet and MS-Restormer outperform current image-denoising methods on all tasks. After adding proposed the multi-sensor consistency supervision, the model can effectively learn the sensor-invariant representations, thereby boosting the generalization capabilities to new sensors.

The results on MSSIDD under the Raw2RGB setting are presented in Table \ref{tab:results_rgb}. Our two models, MS-NAFNet and MS-Restormer also exceed existing works with a notable margin, demonstrating the effectiveness of our method. It should be noted that our method can be easily plugged into existing denoising frameworks while bringing about negligible computational overhead.

\textbf{Visualization.} 
Fig.~\ref{fig_results} provides qualitative visualization under Raw2Raw setting. The denoised images by our method contain more fine-grained details, while the baseline method generate blurred edges or artifacts in texture-rich areas. For instance, in the last row of Fig.~\ref{fig_results}, our model can restore the detailed edge of the text pleasantly, while two baseline methods fail to restore the text clearly.

\subsection{Ablation Study} 
\label{subsec_ablation}
We analyze the effect of sensor consistency supervision on the ``$\rightarrow$ sensor6'' task under Raw2Raw setting, as shown in Fig.~\ref{tab_fig: analysis} (right).
The configuration without $\mathcal{L}_{MS}$ supervision (1st row) serves as the baseline.
Compared to the baseline, each individual part, i.e., $\mathcal{L}_{intra}$, $\mathcal{L}_{inter}$, $\mathcal{L}_{adv}$, can bring stable improvement in terms of PSNR and SSIM. We can observe that the combination of these supervisions (5th row) achieves superior performance, illustrating their effectiveness and complementarity.

Fig.~\ref{tab_fig: analysis} also shows the feature similarity matrix cross sensors before and after applying the sensor consistency supervision $\mathcal{L}_{MS}$, where each pixel denotes the relevance between averaged features from two related sensors. After adding sensor consistency supervision, the relevance among these sensors is more consistent, showing its effectiveness in facilitating features cross-sensor invariance.

\subsection{Results on Real Sensor}

Benefit from the inverse transformation, we can generate raw images using sRGB without the need for image collection workload. We also conduct experiments using raw images captured by real cameras to further validate the effectiveness of the proposed supervision. We employ two additional cameras, i.e., Sony, and Canon, to capture the same scene, and obtain 2000 raw images for each sensor. We subsequently use Sony sRGB image to perform inverse transformation according to the camera parameters of sensor6, resulting in three domains, namely sensor6, Sony, and Canon. The experimental results of Raw2Raw are shown in Tab.\ref{tab_real_sensor}.

\begin{table}[]
\small
\centering
\scalebox{0.95}{
\begin{tabular}{l|c|c|c}
\hline
PSNR/SSIM    & $\rightarrow$ sensor6   & $\rightarrow$ Sony      & $\rightarrow$ Canon     \\ \hline
NAFNet       & 44.20/0.9684 & 45.59/0.9740 & 46.34/0.9778 \\
MS-NAFNet    & 44.37/0.9702 & 45.88/0.9771 & 46.63/0.9793 \\ \hline
Restormer    & 44.38/0.9692 & 45.73/0.9747 & 46.46/0.9787 \\ 
MS-Restormer & 44.51/0.9707 & 46.01/0.9778 & 46.78/0.9802 \\ \hline
\end{tabular}
}
\caption{Results on sensor6 and real sensors Sony, and Canon.}
\label{tab_real_sensor}
\end{table}

\section{Conclusion and Future Work}
\label{sec_conclusion}

In this paper, we introduce the first raw-domain dataset, MSSIDD, for multi-sensor image denoising task.
The proposed MSSIDD comprises six sensors and the data for each sensor are generated based on the specific camera sensor parameters. We simulate white balance, color correction, camera noise, and other aspects among these camera sensors, and provide visualizations to demonstrate their difference. We advocate that the MSSIDD dataset can serve as a standard benchmark for evaluating the sensor transferability of denoising models. We further introduce a novel multi-sensor consistency method to promote sensor-invariant representations. We evaluate previous denoising arts on our dataset and further validate the effectiveness of our method.

Limited by the space, we have not provided more evaluation on other denoising methods. Meanwhile, we believe that in the mobile terminal scenario, fixing some parameters of the denoising model trained on existing sensors and only training some variable parameters on new sensors can also alleviate the memory limitations. This is a promising direction, which remains exploration in future work.

\newpage
\clearpage

\noindent\textbf{Acknowledgement.} This work was supported by National Science Foundation of China (U20B2072, 61976137). This work was also partly supported by SJTU Medical Engineering Cross Research Grant YG2021ZD18.



\newpage
\clearpage

\section{Supplemental Materials}

\subsection{Datasheet}

We apply datasheets for datasets for dataset documentation and intended uses, where we illustrate the motivation for developing the MSSIDD benchmark, dataset composition, collection process, preprocessing, dataset usage, distribution, and dataset maintenance, which can be seen in Supp.\ref{supp_datasheet}. 

A jupyter notebook with a tutorial to download and read the datasets for usage in Pytorch can be found at \url{https://www.kaggle.com/datasets/sjtuwh/mssidd} or \url{https://github.com/shibin1027/MSSIDD/tree/main/notebooks/MSSIDD_demo.ipynb}.

\subsection{Reproducibility of the Baselines}

Considering the scalability of our code, we employ the Basicsr framework and implement the load and usage of MSSIDD dataset to evaluate the sensor transferability of previous methods. Our code is open-source on Github platform \url{https://github.com/shibin1027/MSSIDD}.

\noindent\textbf{Evaluation Metric} Since the raw images in our MSSIDD benchmark are obtained through the inverse ISP transformation and is relatively dark compared with the sRGB images, directly calculating the PSNR and SSIM metrics on raw images can not accurately distinguish the sensor transfer ability of different methods. Therefore, under the RAW2RAW setting, we will conduct the digital gain and gamma correction on the predicted clean raw images output by the network, so that the PSNR and SSIM metrics can be distinguished well across different methods. We set the digital gain here as 2.0 and gamma correction parameter as $1.0/2.2$.

It should be noted that this post-process does not affect the distinction between the settings of RAW2RAW and RAW2RGB. Under the RAW2RAW setting, we apply the supervision in the raw domain, while under the RAW2RGB setting, we apply the supervision in the sRGB domain. For RAW2RGB setting, the raw images output by the network will undergo a complete ISP post-processing to obtain sRGB images and then we calculate loss calculation on sRGB images. We also encourage subsequent methods to adopt this processing to align our baseline.

\subsection{Background and Details of ISP Pipeline}

\textbf{Demosaicing.} The raw images are aranged by Bayer pattern, where in a 2$\times$2 pixel unit, there is one red component pixel, one blue component pixel, and two green component pixels, typically arranged in either RGGB or BGGR patterns. Demosaicing algorithms are applied to produce a full-resolution RGB image, while mosaicing is its inverse process. For the red and blue components, we simply down-sample the corresponding components of the RGB image. For the two green components, we perform two separate down-samplings of the green component in the RGB image at different positions.

\textbf{Digital gain.} To ensure diversity in our dataset, we randomly sample the inverse gain value for each image, where we assume the inverse gain follows a Gaussian distribution with 0.65 mean and 0.2 standard deviation.

\textbf{White balance.} White balance is the process of removing unrealistic color casts so that objects that appear white in person are rendered white in the photo.  Appropriate white balance in cameras takes into account the color temperature of the light source, which affects the coolness or warmth of the perceived white light. While our eyes are adept at recognizing white under various lighting conditions, digital cameras often struggle with auto white balance (AWB).

\section{Datasheet for the MSSIDD Benchmark}
\label{supp_datasheet}

\datasheetsection{Motivation}
\begin{datasheetitem}{For what purpose was the dataset created? \normalfont Was there a specific task in mind? Was there a specific gap that needed to be filled? Please provide a description.}
    MSSIDD is created to be the first publicly available multi-sensor denoising benchmark for evaluating the sensor transferability of the denoising model. The current denoising dataset may generate data from multiple cameras, but they have little difference and are not well partitioned to facilitate multi-sensor transferability evaluation. In a broader sense, the datasets of MSSIDD were also created for unsupervised and semi-supervised learning and partially for supervised learning. 
\end{datasheetitem}
\begin{datasheetitem}{Who created the dataset (e.g., which team, research group) and on behalf of which entity (e.g., company, institution, organization)?}
    As released on May 27, 2024, the initial version of MSSIDD was created by Shibin Mei, Hang Wang, and Bingbing Ni from Huawei Technologies Co., Ltd. and Shanghai Jiao Tong University.
\end{datasheetitem}
\begin{datasheetitem}{Who funded the creation of the dataset? \normalfont If there is an associated grant, please provide the name of the grantor and the grant name and number.}
    There is no specific grant for the creation of the MSSIDD Benchmark. The datasets were created as part of the work at Huawei Technologies Co., Ltd. and Shanghai Jiao Tong University.
\end{datasheetitem}

\datasheetsection{Composition}	
\begin{datasheetitem}{What do the instances that comprise the dataset represent (e.g., documents, photos, people, countries)? \normalfont Are there multiple types of instances (e.g., movies, users, and ratings; people and interactions between them; nodes and edges)? Please provide a description.}
    The instances are raw images used for image-denoising model training. The MSSIDD includes image pairs of noisy raw images and their corresponding clean raw images (ground truth), along with the white balance parameters and color correction matrices of each raw image as metadata.
\end{datasheetitem}
\begin{datasheetitem}{How many instances are there in total (of each type, if appropriate)?}
    \begin{table}[b]
        \caption{Dataset overview. Our dataset includes six sensors and is accompanied by inverse transformation parameters (white balance, color correction) as metadata.}
        \label{tab_supp_dataset_overview}
        \small
        \centering
        \begin{tabular}{lccc}
        \toprule
        Sensor  & total images & train & val  \\ \hline
        sensor1 & 11000        & 10000 & 1000 \\
        sensor2 & 11000        & 10000 & 1000 \\
        sensor3 & 11000        & 10000 & 1000 \\
        sensor4 & 11000        & 10000 & 1000 \\
        sensor5 & 11000        & 10000 & 1000 \\
        sensor6 & 11000        & 10000 & 1000 \\ \bottomrule
        \end{tabular}
    \end{table}
    \autoref{tab_supp_dataset_overview} shows the dataset overview of our MSSIDD benchmark. The dataset consists of six sensors. Each sensor possesses 10000 training raw images and 1000 validation raw images, resulting in a total of 66000 raw images. We also provide the inverse transformation parameters (white balance, color correction) as metadata.
\end{datasheetitem}
\begin{datasheetitem}{Does the dataset contain all possible instances or is it a sample (not necessarily random) of instances from a larger set? \normalfont If the dataset is a sample, then what is the larger set? Is the sample representative of the larger set (e.g., geographic coverage)? If so, please describe how this representativeness was validated/verified. If it is not representative of the larger set, please describe why not (e.g., to cover a more diverse range of instances, because instances were withheld or unavailable).}
    The dataset of MSSIDD contains samples of six different sensors, where we simulate the inverse transformation to obtain the noisy and clean raw image according to the corresponding sensor parameters. The datasets are not representative of all camera sensor scenarios, as the distribution of the camera sensors is highly dynamic and diverse. Instead, the motivation is to resemble the variety of different sensors to facilitate the evaluation of the sensor transferability. Therefore, the MSSIDD should be considered as a sensor transferability benchmark, which makes it possible to evaluate denoising methods in development scenarios.
\end{datasheetitem}
\begin{datasheetitem}{What data does each instance consist of? \normalfont “Raw” data (e.g., unprocessed text or images) or features? In either case, please provide a description.}
    Each instance in the train split consists of the following components: 
    \par\textit{(1)} A single $256\times256$ noisy raw image with 4 channels following BGGR bayer pattern.
    \par\textit{(2)} A single $256\times256$ clean raw image with 4 channels following BGGR bayer pattern.
    \par Each instance in the validation split consists of the following components: 
    \par\textit{(1)} A single $128\times128$ noisy raw image with 4 channels following BGGR bayer pattern.
    \par\textit{(2)} A single $128\times128$ clean raw image with 4 channels following BGGR bayer pattern.
    \par There is an extra ``meta$\_$data.pkl'' file, which contains the white balance, digital gain, and color correction parameters in a dictionary format with image names as key.
\end{datasheetitem}
\begin{datasheetitem}{Is there a label or target associated with each instance? \normalfont If so, please provide a description.}
    As described above, the labels of noisy raw images are the clean raw images.
\end{datasheetitem}
\begin{datasheetitem}{Is any information missing from individual instances? \normalfont If so, please provide a description, explaining why this information is missing (e.g., because it was unavailable). This does not include intentionally removed information, but might include, e.g., redacted text.}
    Everything is included. No data is missing.
\end{datasheetitem}
\begin{datasheetitem}{Are relationships between individual instances made explicit (e.g., users’ movie ratings, social network links)? \normalfont If so, please describe how these relationships are made explicit.}
    There are no relationships made explicit between instances. However, some instances are generated by the same sensor and therefore have an implicit relationship.
\end{datasheetitem}
\begin{datasheetitem}{Are there recommended data splits(e.g., training, development/validation, testing)? \normalfont If so, please provide a description of these splits, explaining the rationale behind them.}
    Each sensor is split into training and validation subsets, as shown in \autoref{tab_supp_dataset_overview}. Since the benchmark is used for sensor transferability evaluation, the data splits are rational.
\end{datasheetitem}
\begin{datasheetitem}{Are there any errors, sources of noise, or redundancies in the dataset? \normalfont If so, please provide a description.}
    MSSIDD benchmark is generated from clean sRGB images from the SIDD dataset. We obtain six raw image pairs from one source sRGB image, which results in similar scenarios but conveys distinct color and noise levels. We have double-checked our dataset and there are no errors.
\end{datasheetitem}
\begin{datasheetitem}{Is the dataset self-contained, or does it link to or otherwise rely on external resources (e.g., websites, tweets, other datasets)? \normalfont If it links to or relies on external resources, a) are there guarantees that they will exist, and remain constant, over time; b) are there official archival versions of the complete dataset (i.e., including the external resources as they existed at the time the dataset was created); c) are there any restrictions (e.g., licenses, fees) associated with any of the external resources that might apply to a dataset consumer? Please provide descriptions of all external resources and any restrictions associated with them, as well as links or other access points, as appropriate.}
    MSSIDD is entirely self-contained.
\end{datasheetitem}
\begin{datasheetitem}{Does the dataset contain data that might be considered confidential (e.g., data that is protected by legal privilege or by doctor-patient confidentiality, data that includes the content of individuals’ non-public communications)? \normalfont If so, please provide a description.} 
    The MSSIDD dataset does not contain data that might be considered confidential since they are generated with open-source images and open camera sensor parameters.
\end{datasheetitem}
\begin{datasheetitem}{Does the dataset contain data that, if viewed directly, might be offensive, insulting, threatening, or might otherwise cause anxiety? \normalfont If so, please describe why.}
    No.
\end{datasheetitem}
\begin{datasheetitem}{Does the dataset identify any subpopulations (e.g., by age, gender)? \normalfont If so, please describe how these subpopulations are identified and provide a description of their respective distributions within the dataset.}
    No.
\end{datasheetitem}
\begin{datasheetitem}{Is it possible to identify individuals (i.e., one or more natural persons), either directly or indirectly (i.e., in combination with other data) from the dataset? \normalfont If so, please describe how.}
    No. 
\end{datasheetitem}
\begin{datasheetitem}{Does the dataset contain data that might be considered sensitive in anyway(e.g., data that reveals race or ethnic origins, sexual orientations, religious beliefs, political opinions or union memberships, or locations; financial or health data; biometric or genetic data; forms of government identification, such as social security numbers; criminal history)? \normalfont If so, please provide a description.}
    No.
\end{datasheetitem}

\datasheetsection{Collection Process}	
\begin{datasheetitem}{How was the data associated with each instance acquired? \normalfont Was the data directly observable (e.g., raw text, movie ratings), reported by subjects (e.g., survey responses), or indirectly inferred/derived from other data (e.g., part-of-speech tags, model-based guesses for age or language)? If the data was reported by subjects or indirectly inferred/derived from other data, was the data validated/verified? If so, please describe how.}
    Each instance in MSSIDD is obtain by inverse ISP transformation from original sRGB image in SIDD dataset. Starting from clean sRGB images from the SIDD dataset, we sequentially apply inverse tone mapping, inverse gamma correction, inverse color correction, inverse white balance with digital gain adjustments, and inverse demosaicing (mosaic), with a corresponding sensor parameter, to simulate the clean images in the raw domain, thus obtaining the clean raw image $I_2$. We then add the shot and read noise according to sensor noise calibration parameter to obtain the noisy raw image $I_1$. We can finally get an instance pair denoted as $(I_1, I_2)$.
\end{datasheetitem}
\begin{datasheetitem}{What mechanisms or procedures were used to collect the data (e.g., hardware apparatuses or sensors, manual human curation, software programs, software APIs)? \normalfont How were these mechanisms or procedures validated?}
    The data is generated with inverse ISP transformation, which was first proposed by ``Unprocessing Images for Learned Raw Denoising''. More information can be found in the corresponding paper which has been well validated. 
\end{datasheetitem}
\begin{datasheetitem}{If the dataset is a sample from a larger set, what was the sampling strategy (e.g., deterministic, probabilistic with specific sampling probabilities)?}
    The MSSIDD dataset is not sampled from a larger dataset, but the source sRGB image for inverse ISP are sampled from SIDD with a random sample.
\end{datasheetitem}
\begin{datasheetitem}{Who was involved in the data collection process (e.g., students, crowdworkers, contractors) and how were they compensated (e.g., how much were crowdworkers paid)?}
    Only the authors are involved in the collection process. 
\end{datasheetitem}
\begin{datasheetitem}{Over what timeframe was the data collected? \normalfont Does this timeframe match the creation timeframe of the data associated with the instances (e.g., recent crawl of old news articles)? If not, please describe the timeframe in which the data associated with the instances was created.}
    The MSSIDD dataset was collected and annotated from March 2024 to June 2024. 
\end{datasheetitem}
\begin{datasheetitem}{Were any ethical review processes conducted (e.g., by an institutional review board)? \normalfont If so, please provide a description of these review processes, including the outcomes, as well as a link or other access point to any supporting documentation.}
    No ethical reviews have been conducted to date. However, an ethical review may be conducted as part of the paper review process.
\end{datasheetitem}

\datasheetsection{Preprocessing/cleaning/labeling}	
\begin{datasheetitem}{Was any preprocessing/cleaning/labeling of the data done (e.g., discretization or bucketing, tokenization, part-of-speech tagging, SIFT feature extraction, removal of instances, processing of missing values)? \normalfont If so, please provide a description. If not, you may skip the remaining questions in this section.}
    There is no preprocessing/cleaning/labeling of the data. All raw images are 4 channels with RGGB bayer pattern, and stored in \texttt{numpy.uint16} type.
\end{datasheetitem}
\begin{datasheetitem}{Was the “raw” data saved in addition to the preprocessed/cleaned/labeled data (e.g., to support unanticipated future uses)? \normalfont If so, please provide a link or other access point to the “raw” data.}
    No.
\end{datasheetitem}
\begin{datasheetitem}{Is the software that was used to preprocess/clean/label the data available? \normalfont If so, please provide a link or other access point.}
    N/A. The MSSIDD does not involve preprocess/clean/label software.
\end{datasheetitem}

\datasheetsection{Uses}	
\begin{datasheetitem}{Has the dataset been used for any tasks already? \normalfont If so, please provide a description.}
    The datasets were used to create denoising model sensor transferability baselines for the corresponding paper presenting the MSSIDD Benchmark.
\end{datasheetitem}
\begin{datasheetitem}{Is there a repository that links to any or all papers or systems that use the dataset? \normalfont If so, please provide a link or other access point.}
    Yes, the baselines presented in the corresponding paper ``MSSIDD: A Benchmark for Multi-Sensor Denoising''.
\end{datasheetitem}
\begin{datasheetitem}{What(other) tasks could the dataset be used for?}
    The MSSIDD can be used for sensor transferability evaluation. In a broader sense, the datasets of MSSIDD can also be used for unsupervised and semi-supervised learning and partially for supervised learning.
\end{datasheetitem}
\begin{datasheetitem}{Is there anything about the composition of the dataset or the way it was collected and preprocessed/cleaned/labeled that might impact future uses? \normalfont For example, is there anything that a dataset consumer might need to know to avoid uses that could result in unfair treatment of individuals or groups (e.g., stereotyping, quality of service issues) or other risks or harms (e.g., legal risks, financial harms)? If so, please provide a description. Is there anything a dataset consumer could do to mitigate these risks or harms?}
    No.
\end{datasheetitem}
\begin{datasheetitem}{Are there tasks for which the dataset should not be used? \normalfont If so, please provide a description.}
    No.
\end{datasheetitem}

\datasheetsection{Distribution}	
\begin{datasheetitem}{Will the dataset be distributed to third parties outside of the entity (e.g., company, institution, organization) on behalf of which the dataset was created? \normalfont If so, please provide a description.}
    Yes, MSSIDD is publicly available on the internet for anyone interested in using it. 
\end{datasheetitem}
\begin{datasheetitem}{How will the dataset will be distributed (e.g., tarball on website, API, GitHub)? \normalfont Does the dataset have a digital object identifier (DOI)?}
    MSSIDD is distributed through kaggle at \url{https://www.kaggle.com/datasets/sjtuwh/mssidd}.
    \par DOI: 10.34740/kaggle/dsv/8533876
\end{datasheetitem}
\begin{datasheetitem}{When will the dataset be distributed?}
    The datasets have been available on kaggle since May 28, 2024.
\end{datasheetitem}
\begin{datasheetitem}{Will the dataset be distributed under a copyright or other intellectual property (IP) license, and/or under applicable terms o fuse (ToU)? \normalfont If so, please describe this license and/or ToU, and provide a link or other access point to, or otherwise reproduce, any relevant licensing terms or ToU, as well as any fees associated with these restrictions.}
    MSSIDD is licensed under the Apache License Version 2.0, January 2004.
\end{datasheetitem}
\begin{datasheetitem}{Have any third parties imposed IP-based or other restrictions on the data associated with the instances? \normalfont If so, please describe these restrictions, and provide a link or other access point to, or otherwise reproduce, any relevant licensing terms, as well as any fees associated with these restrictions.}
    No.
\end{datasheetitem}
\begin{datasheetitem}{Do any export controls or other regulatory restrictions apply to the dataset or to individual instances? \normalfont If so, please describe these restrictions, and provide a link or other access point to, or otherwise reproduce, any supporting documentation.}
    Unknown to authors of the datasheet.
\end{datasheetitem}

\datasheetsection{Maintenance}	
\begin{datasheetitem}{Who will be supporting/hosting/maintaining the dataset?}
    MSSIDD is hosted on kaggle and supported and maintained by the authors.
\end{datasheetitem}
\begin{datasheetitem}{How can the owner/curator/manager of the dataset be contacted (e.g., email address)?}
    The curators of the datasets can be contacted under shibin.mei1027@gmail.com or francis970625@gmail.com.
\end{datasheetitem}
\begin{datasheetitem}{Is there an erratum? \normalfont If so, please provide a link or other access point.}
    No. 
\end{datasheetitem}
\begin{datasheetitem}{Will the dataset be updated (e.g., to correct labeling errors, add new instances, delete instances)? \normalfont If so, please describe how often, by whom, and how updates will be communicated to dataset consumers (e.g., mailing list, GitHub)?}
    New versions of MSSIDD datasets will be shared at kaggle if corrections are necessary.
\end{datasheetitem}
\begin{datasheetitem}{Will older versions of the dataset continue to be supported/hosted/maintained? \normalfont If so, please describe how. If not, please describe how its obsolescence will be communicated to dataset consumers.}
    Yes, we plan to support versioning of the datasets so that all the versions are available to potential users. We maintain the history of versions at kaggle. Each version will have a unique DOI assigned.
\end{datasheetitem}
\begin{datasheetitem}{If others want to extend/augment/build on/contribute to the dataset, is there a mechanism for them to do so? \normalfont If so, please provide a description. Will these contributions be validated/verified? If so, please describe how. If not, why not? Is there a process for communicating/distributing these contributions to dataset consumers? If so, please provide a description.}
    Others can extend/augment/build on MSSIDD by contacting us to obtain administration for the dataset. Others are free to release their extension of the MSSIDD Benchmark or its datasets under the Apache License Version 2.0, January 2004.
    
\end{datasheetitem}

\end{document}